# Real-Time Shape Tracking of Facial Landmarks*

Hyungjoon Kim
School of Electronic Engineering
Korea University
Republic of Korea
hyungjun89@korea.ac.kr

Hyeonwoo Kim
School of Electronic Engineering
Korea University
Republic of Korea
guihon12@korea.ac.kr

Eenjun Hwang
School of Electronic Engineering
Korea University
Republic of Korea
ehwang04@korea.ac.kr

## ABSTRACT

Detection of facial landmarks and accurate tracking of their shape are essential in real-time virtual makeup applications, where users can see the makeup's effect by moving their face in different directions. Typical face tracking techniques detect diverse facial landmarks and track them using a point tracker such as the Kanade-Lucas-Tomasi (KLT) point tracker. Typically, 5 or 64 points are used for tracking a face. Even though these points are sufficient to track the approximate locations of facial landmarks, they are not sufficient to track the exact shape of facial landmarks. In this paper, we propose a method that can track the exact shape of facial landmarks in real-time by combining a deep learning technique and a point tracker. We detect facial landmarks accurately using SegNet, which performs semantic segmentation based on deep learning. Edge points of detected landmarks are tracked using the KLT point tracker. In spite of its popularity, the KLT point tracker suffers from the point loss problem. We solve this problem by executing SegNet periodically to calculate the shape of facial landmarks. That is, by combining the two techniques, we can avoid the computational overhead of SegNet for real-time shape tracking and the point loss problem of the KLT point tracker. We performed several experiments to evaluate the performance of our method and report some of the results herein.

## 1 INTRODUCTION

With the advancement of information technology, many new applications have become feasible. For instance, in virtual makeup, people can verify the result of applying certain makeup on their face by selecting cosmetic products and makeup style. This kind of personalized experience is useful in purchasing cosmetic products or determining makeup style. Furthermore, with the increasing popularity of online shopping, virtual experience has become more critical than ever as it can contribute to purchasing relevant products.

Recently, several smartphone applications have been developed to present various effects to a person's face using decoration and makeup. These applications recognize the user's face, detect its facial landmarks, and synthesize each landmark with a predefined template. Therefore, such synthesis usually requires a fixed view of a face, such as front view or side view depending on the type of decoration or makeup. This has been a severe restriction because users want to see the 3D effect of such decoration or makeup by moving their face in different directions.

This restriction can be alleviated if we can detect facial landmarks and perform the decoration or makeup in real-time. This is especially useful to verify the effect of a particular cosmetic product.

Using this technique, we can effectively change the makeup of a specific character in a video. To do this, we need to track facial landmarks in real-time. Thus far, face-tracking techniques have calculated facial landmark points using image processing, machine learning, or deep learning, and have tracked those points using a point tracker. Typically, 5 or 64 facial landmark points are used; facial landmarks do not include the hair region. Since these points are sufficient to track the approximate location of facial landmarks, they can be used to synthesize a template based on the original image. However, for virtual makeup and sophisticated face tracking, we need to track the exact shape of the facial landmarks in addition to their location.

On the other hand, most face-tracking techniques do not consider hair despite it being an important component of makeup that needs to be tracked. The problem with hair is that it is difficult to detect its shape because of the wide range of colors and shapes in the real world. In [1], authors proposed a method that determines personal color by analyzing a face image and performs virtual makeup according to the personal color. In order to detect facial landmarks, they used Dlib and applied makeup to the detected landmarks based on their coordinates. Results showed that the virtual makeup was not good because the coordinates did not represent the landmarks accurately. In addition, the method showed limited performance in detecting hair region and thus hair coloring could not be performed effectively. To solve these problems, they proposed a new facial landmark detection method based on SegNet and showed that virtual makeup, including the hair region, can be performed effectively [2].

As previously mentioned, for a more effective virtual makeup system, users need to observe the makeup effect as they move their face. SegNet is a deep learning technique used for classifying a large-scale image repository and requires significant computational resources. In our experiment, the processing time of SegNet for landmark detection was approximately 0.15 s. This is enough for processing a single image, but it is too slow to handle video stream that requires a processing speed of 30 fps. To solve this problem, we use SegNet to detect facial landmarks of a video frame and then track the landmarks using the Kanade-Lucas-Tomasi (KLT) point tracker. Even though this method is fast, it suffers from the point loss problem. This problem can be solved by executing SegNet periodically and using the detected landmarks for tracking.

This paper is organized as follows; Section 2 introduces related works and Section 3 describes how to track the exact shape using facial landmarks in real-time. Experimental results for the proposed method are described in Section 4 and Section 5 concludes the paper.

## 2 Related Works

Several past works focused on recognizing face in an image and tracking it in real-time. Recently, Convolutional Neural Network (CNN) has significantly improved the performance of image classification and recognition, especially on a large scale. Real-time face tracking requires quick and accurate facial landmark recognition. Depending on the purpose, exact shape and location of facial landmarks should be detected and tracked.

### 2.1 Face Tracking

Face tracking is a technology that can be used in various fields. During the tracking process, as we have to predict the position of the next frame features, the change of illumination, facial expression, and position becomes a problem. Various tracking techniques have been developed to address this. Zaheer Shaik and Vijayan Asari proposed face detection and tracking method [3]. Face features are obtained using Viola–Jones face detection [4] in the first frame and then the Kalman filter [5] is used for predicting the position of the points in next frame. In [6], feature points were extracted using the KLT point tracker and then facial tracking was performed using the pyramidal Lucas-Kanade Feature Tracker. Wettum performed a comparative experiment to determine the best algorithm for real-time tracking with smart phone [7]. To compare the four algorithms, namely Lucas-Kanade (LK) [8] point tracker, Structured Output Tracking with Kernels (Struck) [9], Discriminative Scale Space Tracker (DSST) [10], and Kernelized Correlation Filters (KCF) [11], the Dlib Facial Landmark Detector (DFLD), which is a facial landmark localization library 12] and Deformable Shape Tracking (DEST) [13] were used as comparison objects. Results indicate that LK tracker and DSST are the algorithms that can be used for actual facial landmark tracking. However, as DSK cannot be performed in real-time, we can conclude that LK is the most useful algorithm for facial landmark tracking. In some cases, deep learning has helped detect objects in real-time. Kaipeng Zhang et al. proposed a lightweight CNN model called multi-task cascaded convolutional networks (MTCNN) [14] for real-time face detection. The MTCNN consists of three CNNs—P-Net, R-Net, and O-Net. In this process, the candidate bounding boxes are first produced to P-Net, then the refined bounding boxes are provided to the R-NET, and finally the O-Net produces final bounding box and facial landmark position.

### 2.2 Object Recognition

Ross Girshick proposed R-CNN for object detection [15]. R-CNN uses selective search algorithm that combines adjacent pixels with similar color or intensity patterns to find the bounding box, after which it resizes the extracted bounding box to input to CNN, and finally sorts the image using Support Vector Machine (SVM). However, R-CNN has the disadvantage of low speed because all the bounding boxes are fed to CNN as input. To solve this problem, Shaoqing Ren et al. proposed Fast R-CNN [16] that uses the concept of region of interest pooling (RoIPooling). They collected RoI information using selective search, and constructed a feature map for the entire input image using CNN. Then, they extracted a bounding box area by evaluating the stored RoI information using RoIPooling. Still, selective search is time-consuming. Shaoqing Ren et al. proposed Faster R-CNN [17] that contains Region Proposal Network (RPN). RPN is a CNN that plays a role similar to selective search to create RoI. Another method for real-time object detection is the You Only Look Once (YOLO) proposed by Joseph Redmon et al [18]. YOLO divides an image into a N x N grid and for each grid, it calculates B bounding boxes, confidence for those bounding boxes, and C class probabilities. The confidence score of a box indicates its accuracy in recognizing objects in the bounding box. Final bounding boxes are obtained by combining the confidence score and class probability map.

### 2.3 Semantic Segmentation

Jonathan Long et al. first proposed a model for semantic segmentation [19] called Fully Convolutional Networks (FCN) that replaced the fully connected layer of CNN with a convolutional layer. They then used a method to restore the pixel information by upsampling the output with the features of the previous layer. However, the output from the convolutional layer and the pooling layer has a resolution problem. Moreover, the details also disappear because this resolution is upsampled again. To solve this problem, So Noh. H et al. proposed deconvNet [20]. DeconvNet consists of a convolutional network and its corresponding deconvolutional network. After storing the pooling location information in the convolutional network, it is used for unpooling in the deconvolutional network. Therefore, more sophisticated restoration is possible, and the details of pixel information can be saved. Similarly, Olaf Ronneberger et al. proposed U-net [21]. U-net is used for upconvolution by copying and cropping the feature map from the convolution layer. Vijay Badrinarayanan et al. proposed SegNet [22], which is a combination of deconvNet and U-net structures. The encoder is composed of several CNN layers, and the decoder network has a corresponding structure. Unlike deconvNet, SegNet reduces parameterization that requires computational resources and adds pooling indices that are not in U-net for unpooling.

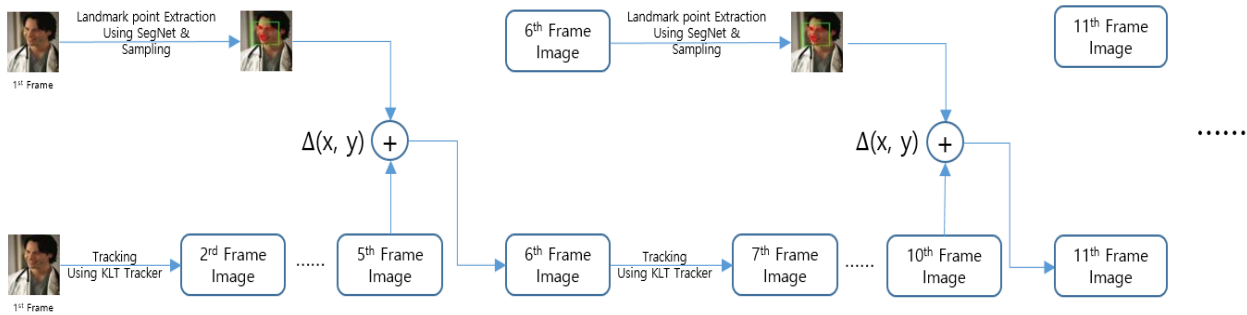

**Figure 1:** Overall architecture for semantic facial landmark tracking

## 3 Proposed Method

Although computer hardware has been steadily improving, it is still difficult and expensive to perform real-time semantic analysis for video stream based on deep learning. Thus, in this paper, we combine a deep learning technique for exact landmark detection and a traditional tracking technique for fast landmark tracking. Figure 1 shows our semantic tracking process. First, we detect facial landmarks using SegNet. Then, we calculate edge points of facial landmarks using Canny edge detection and track those points using the KLT point tracker. If landmark points disappear for certain reasons during tracking, the KLT point tracker may lose the points and its tracking result remains incomplete. If face movement is too fast, the tracked points may not locate the landmark correctly. Therefore, we use the outcome of SegNet and Canny edge detection periodically to refresh the tracking points of the tracker. In this way, we overcome the computational overhead of deep learning and the point loss problem of the KLT point tracker.

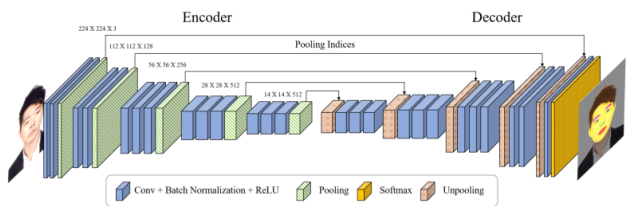

**Figure 2:** SegNet architecture

### 3.1 Facial Landmark Points Extraction

As described in earlier sections, we used SegNet to detect facial landmarks. The basic network structure is almost the same as that used in [22] and its overall architecture is shown in Figure 2. We maintained the initial settings in [22] and trained the network using pre-trained VGG16 model, which is a deep learning technique designed to classify an ILSVRC dataset. Initial training dataset consists of 150 frontal face images and 30 profile face images. In order to strengthen our method, we produced and added more varied images in terms of color and shape by performing diverse operations such as image rotation and noise filtering on the original images. As a result, we obtained a total of 6840 images as training set. We defined nine classes to describe facial landmark features and produced ground truth manually for each image in the dataset. Figure 3 shows some of the original images and their transformed images created using rotation, noise insertion, or both. Figure 4 shows the ground truth for each image in Figure 3. In this figure, each landmark is marked with a different color defined in Table 1.

Table 1: Facial Landmark and R, G, B Triple

|  | R | G | B | Color |
| --- | --- | --- | --- | --- |
| Hair | 106 | 57 | 6 | |
| Face skin | 255 | 255 | 0 | |
| Sclera | 0 | 255 | 0 | |
| Pupil | 0 | 0 | 255 | |
| Eyebrow | 255 | 0 | 255 | |
| Nostril | 0 | 255 | 255 | |
| Lip | 255 | 0 | 0 | |
| Between mouth | 255 | 255 | 255 | |
| Background | 0 | 0 | 0 | |

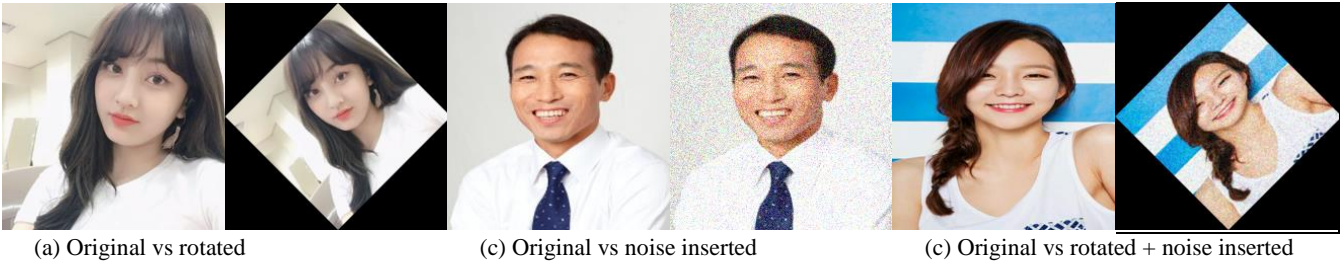

(a) Original vs rotated    (c) Original vs noise inserted    (c) Original vs rotated + noise inserted

**Figure 3:** **Construction of dataset**

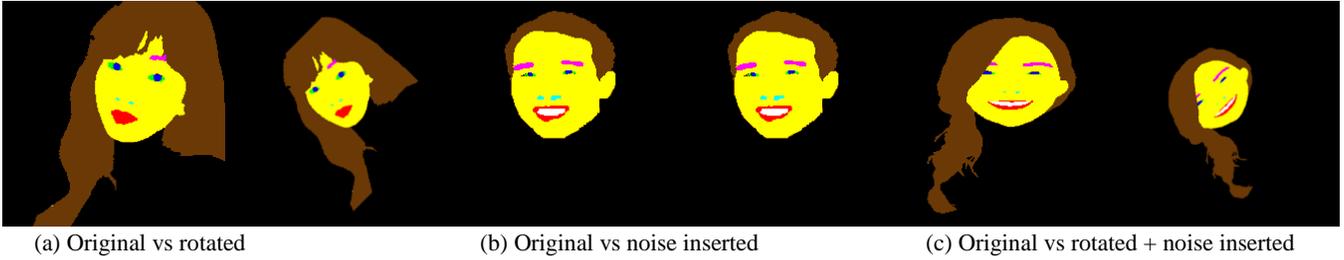

(a) Original vs rotated    (b) Original vs noise inserted    (c) Original vs rotated + noise inserted

**Figure 4:** **Ground truth by different color**

Figure 4 shows that the transformed images have the same ground truth as the original images. This is because the ground truth reflects only the shape of the landmark, and the noise does not affect the shape of object but it does affect intensity or color in the image.

Facial landmarks are extracted using SegNet, and the mask of each landmark is created for semantic tracking. The mask is a set of points that represent each facial landmark region. We use these points for tracking. However, landmarks consist of thousands, even tens of thousands, of pixels. Thus, tracking all the points using the KLT point tracker requires significant computational time and cannot maintain the video stream speed. To reduce the number of points to be tracked without degrading the landmark shape, we need to perform point sampling that satisfies the following requirements:

1. The number of points should be small enough to meet the required FPS.
2. It should be easy to recover the landmark shape from the sampled points.

As a point sampling to satisfy these conditions, we decide to utilize the contour points of a landmark as tracking points of the landmark. To obtain the contour, we first apply the median filter [23] on the facial landmark mask to remove noises and then we apply Canny edge detection [24]. The median filter is known to be effective on salt and pepper noise, which is similar to the noise during the landmark detection by SegNet. Figure 5 shows an example of point sampling process. Figure 5 (a) is a facial landmark mask in grayscale image, and Figures 5 (b) and (c) represent the result of applied median filter on each landmark and the contour of the facial landmark, respectively. We use white points in Figure 5 (c) as facial landmark points.

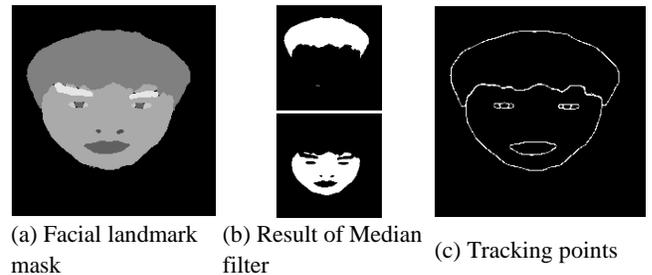

(a) Facial landmark mask    (b) Result of Median filter    (c) Tracking points

**Figure 5:** **Calculating facial landmark points for tracking**

## 3.2 Kanade-Lucas-Tomasi Point Tracker

To track the landmark points, we used the Kanade-Lucas-Tomasi (KLT) point tracker [25]. The key premise of the KLT point tracker is that during a very short period, there is no change in brightness and only a change in position occurs. Let the pixel brightness of arbitrary coordinates x and y at time t be I (x, y, t). If $\delta x$ and $\delta y$ are the distance moved during $\delta t$, which is very short, then the brightness can be represented by $I(x+\delta x, y+\delta y, t+\delta t)$. According to this premise, $I(x, y, t) = I(x+\delta x, y+\delta y, t+\delta t)$. In terms of face video, the difference between two adjacent frames is not substantial. That is, $\delta x$ and $\delta y$ are very small. Figure 6 shows an example of the difference between two adjacent frames. Figure 6 (a) shows two adjacent frames in a video and Figure 6 (b) shows their difference. The difference image is reversed. Colored pixels represent the difference between two frames and each white pixel indicates that there is no difference between two frames at that point.

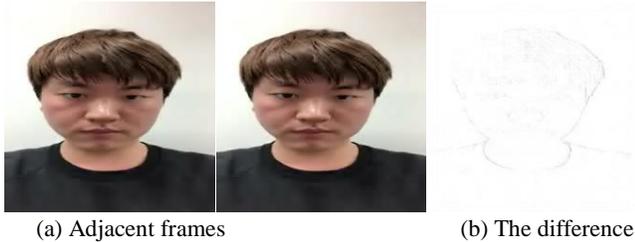

(a) Adjacent frames  (b) The difference

**Figure 6:** **Difference between two adjacent frames**

As shown in Figure 6, typical face video satisfies the premise of the KLT algorithm. Therefore, it is reasonable to apply the KLT algorithm for face tracking. Another advantage of the KLT algorithm is that it is fast enough to conduct real-time processing. For instance, in [25], they demonstrated point tracking using the KLT algorithm with GPU in 2008. Based on the hardware performance at that time, they achieved processing speed of over 200 FPS for 720 × 576 resolution video. Current hardware performance allows real-time tracking even without GPU. Regardless of these advantages of the KLT point tracker, it has several limitations. First, if the tracked object disappears for some reason, such as if the object is out of screen, the tracking points are lost. Second, if the points become completely different, such as if the face is turned to both sides, point loss problem may occur or it will not describe the facial landmark correctly even when the number of points is maintained. These lost points are difficult to recover. Figure 7 shows an example of the point loss problem.

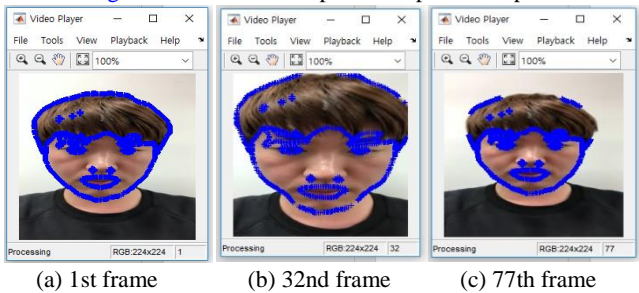

(a) 1st frame  (b) 32nd frame  (c) 77th frame

**Figure 7:** **An example of the point loss problem**

Figure 7 (a) shows the first frame that has initial facial landmark points calculated by SegNet and point sampling. Figure 7 (b) shows the 32nd frame, which is one second later. In this frame, a part of the participant's hair goes beyond the screen while moving his face closer to the camera. Figure 7 (c) shows the 77th frame where the lost points are not recovered even when the participant moves his face away from the camera.

### 3.3 Combined Point Tracker

In the previous section, we showed that the KLT algorithm is not appropriate for real-time semantic tracking because of several limitations. We thus proposed to use the KLT point tracker and SegNet where we first detected facial landmarks from the first frame of face video using SegNet and then tracked them using the KLT algorithm. Average processing speed of the SegNet in our experiment was approximately 0.12 s. To maintain 30 FPS of face video, the processing speed should be higher than 34 FPS. Therefore, when tracking using the KLT point tracker, SegNet performs landmark detection every four frames simultaneously. When the feature detection is complete, the current track points used in the KLT algorithm are replaced with the new ones. SegNet points are more accurate than the current KLT tracker's points. However, considering the difference between the four frames, better precision in tracking can be achieved. Thus, we add an average variation of last four frames on SegNet points.

$$P_C(x,y,t) = P_S(x,y,t) + \Delta(x,y) \quad (1)$$

Here, $P_c$ is a set of modified points in time $t$ that are used in the KTL point tracker and $P_s$ is a set pf points that are obtained from SegNet in time $t$. $\Delta(x,y)$ is defined as

$$\Delta(x,y) = \frac{1}{3}\sum_{i=0}^{2}\frac{P_K(x,y,t-i) - P_K(x,y,t-i-1)}{n_i} \quad (2)$$

$P_K$ is a set of points that are calculated from the KLT algorithm in time $t$ and $n$ is the number of total points. Note that four frames take approximately 0.12 s, which is too unsubstantial to cause the point loss problem.

## 4 Experiment

To evaluate the performance of our proposed method, we performed three experiments. First, we evaluate how accurately SegNet detects facial landmarks. Next, we compare the processing time and FPS of our method, SegNet, and the KLT point tracker. Finally, we evaluate the accuracy of facial landmark tracking when using both SegNet and the KLT point tracker. We performed the experiment under The Intel (R) Core (TM) i7-7700k CPU, 32GB DDR4 RAM, and NVIDIA Geforce GTX 1080Ti, and the development tools were Visual Studio 2015, Matlab 2017b, and Python 3.5.

### 4.1 Landmark Points Detection Accuracy

To evaluate the landmark detection accuracy of SegNet, we used 100 face images not included in the training set. We created the ground truth for each image and evaluated the landmark detection result by comparing it with the correct answer set. Figure 8 shows the landmark detection result. In the figure, (a) and (b) represent input face images and their landmark detection result overlaid with the original images, respectively. As shown in the figure, every landmark is well detected, and their accuracy is shown in Table 2. The accuracy was calculated through pixel-to-pixel matching. When performing pixel-unit matching, we set the threshold value to zero, which indicates that the detected landmark is exactly matched to the predefined answer of the landmark.

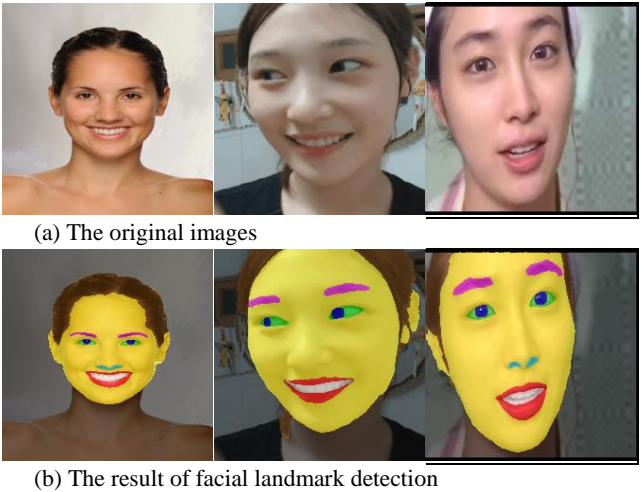

(a) The original images

(b) The result of facial landmark detection

**Figure 8:** **Facial landmark detection result**

**Table 2: Facial Landmark Extraction Accuracy**

| | BACKGROUND | HAIR | EYEBROW | SCLERA | PUPIL | NOSTRIL | MOUTH | INNER MOUTH | FACE SKIN |
|---|---|---|---|---|---|---|---|---|---|
| BACKGROUND | 0.9942 | 0.0034 | 0 | 0 | 0 | 0 | 0 | 0 | 0.0019 |
| HAIR | 0.0122 | 0.9429 | 0 | 0.0002 | 0.0011 | 0 | 0 | 0 | 0.0436 |
| EYEBROW | 0 | 0 | 0.9437 | 0 | 0 | 0 | 0 | 0 | 0.0563 |
| SCLERA | 0 | 0 | 0 | 0.9692 | 0.0308 | 0 | 0 | 0 | 0 |
| PUPIL | 0 | 0 | 0 | 0.0290 | 0.9710 | 0 | 0 | 0 | 0 |
| NOSTRIL | 0 | 0 | 0 | 0 | 0 | 0.9891 | 0 | 0 | 0.0109 |
| MOUTH | 0 | 0 | 0 | 0 | 0 | 0 | 0.9286 | 0.0621 | 0.0093 |
| INNER MOUTH | 0 | 0 | 0 | 0 | 0 | 0 | 0.0032 | 0.9852 | 0.0116 |
| FACE SKIN | 0.0016 | 0.0130 | 0.0130 | 0.0083 | 0.0016 | 0.0107 | 0.0094 | 0.0003 | 0.9942 |

As shown in the table, most landmarks are well detected with an accuracy higher than 90%, and majority of the errors occur for neighboring face region. For example, the pupil could be mistaken as sclera, or the skin could be mistaken as hair or background. As we have confirmed that SegNet can accurately detect facial landmarks, we use the landmarks extracted from each frame using SegNet as the ground truth when evaluating tracking accuracy in the following experiments.

## 4.2 Evaluation of Frame per Second

We propose a real-time semantic tracking technique that combines SegNet and the KLT algorithm. Even though SegNet is accurate, it is relatively slow. On the other hand, the KLT algorithm is fast but less stable. The most important factor in real-time tracking is the processing speed, which should be fast enough to maintain the FPS of the video stream. In order to evaluate real-time tracking of our method, we used five videos of approximately 15 seconds each. The videos used in the experiment are resized to 224 x 224 to apply SegNet and have a frame rate of 30 FPS. Sample frames of the video used in the experiment are shown in Fig. 9.

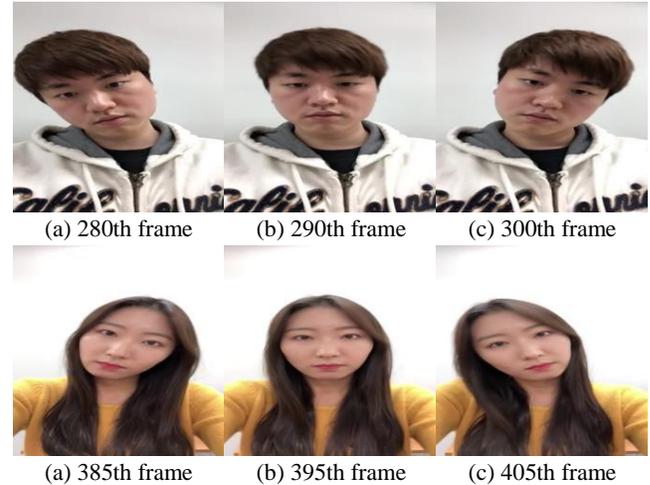

(a) 280th frame  (b) 290th frame  (c) 300th frame

(a) 385th frame  (b) 395th frame  (c) 405th frame

**Figure 9:** **Sample frames of videos**

To measure how many frames can be processed per second, we calculated the processing time of the KLT algorithm and SegNet for one frame. The processing time of the KLT algorithm indicates the time taken to calculate the coordinates of tracking points in the next frame for the tracking points in the current frame. In the case of SegNet, it represents the time taken to detect facial landmarks from the input frame and perform point sampling. We measured the FPS approximately 2500 times and calculated their average processing time and maximum possible FPS, which are shown in Table 3.

Hence, to maintain 30 FPS, which is the play rate of a typical video stream, approximately 34 images per second should be processed. While the KLT algorithm is fast enough, SegNet requires approximately 0.12 second to process a single frame, which corresponds to three to four frames. Accordingly, accurate detection of facial landmarks after every four frames using SegNet and shape tracking by the KLT algorithm accomplishes our goal without the need for any expensive hardware. The FPSs of SegNet, the KLT algorithm, and their combination are shown in Table 4.

**Table 3:** Processing time and Max FPS

| Method | Processing time | Max. FPS |
|---|---|---|
| KLT | 0.0052 s | 192 FPS |
| SegNet | 0.122 s | 8.2 FPS |

**Table 4:** Frame per Second

| Method | FPS |
|--------|-----|
| KLT | 30 |
| SegNet | 7 |
| Combined | 30 |

## 4.3 Landmark Points Tracking Accuracy

In the first experiment, we demonstrated that SegNet can detect facial landmarks accurately. Hence, we utilized SegNet to detect landmarks for all the frames and use them as the ground truth to evaluate how well the points tracked by the KLT algorithm are matched with the ground truth. At this time, the threshold value is three. This indicates that if the difference between the ground truth and the detected landmarks is less than or equal to three pixels, the tracking is considered to be correct. For all the frames in the videos, we evaluated this accuracy. Figure 10 shows the results.

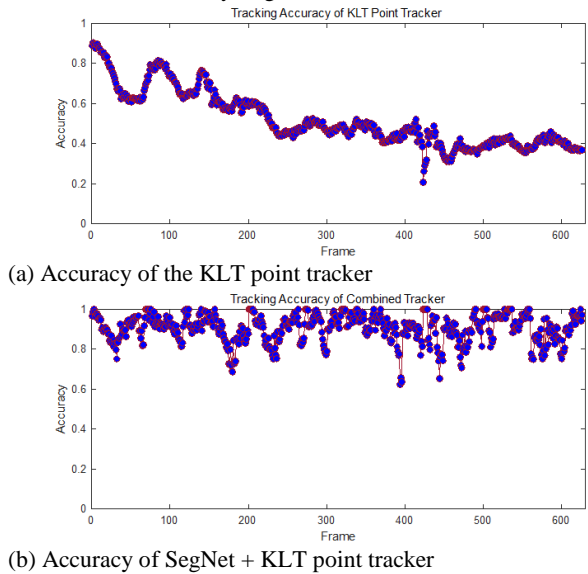

(a) Accuracy of the KLT point tracker

(b) Accuracy of SegNet + KLT point tracker

**Figure 10: Accuracy of facial landmark points tracking**

Figures 10 (a) and (b) show the accuracy of the KLT point tracker and SegNet + KLT point tracker, respectively. As shown in the figures, the accuracy of the KLT point tracker is gradually decreasing because of the persistent point loss problem. On the other hand, our method maintained the accuracy between 80% and 100%. The average accuracy of the KLT point tracker was 52% and that of our method was 86%.

## 5 Conclusion

In this paper, we proposed a scheme for real-time semantic tracking of facial landmarks for video stream. To that end, we utilized the advantages of SegNet and the KLT point tracker while compensating for their weaknesses. That is, we performed accurate facial landmark detection using SegNet and tracked the detected landmark shape swiftly using the KLT point tracker. To evaluate the performance of our method, we performed extensive experiments that showed that the proposed method can accurately perform facial landmark tracking for a typical video stream. We expect that this method can be used in various fields of image processing that require real-time tracking while maintaining the exact shape of the object, such as in virtual real-time makeup or AR-based real-time image synthesis applications. We are currently developing a virtual makeup system based on the proposed method.

## REFERENCES

[1] J. Park, H. Kim, S. Ji, and E. Hwang. 2018. An Automatic Virtual Makeup Scheme Based on Personal Color Analysis. International Conference on Ubiquitous Information Management and Communication (IMCOM 2018).

[2] H. Kim, J. Park, H. Kim, and E. Hwang. 2018. Facial Landmark Extraction Scheme Based on Semantic Segmentation. 2018 International Conference on Platform Technology and Service (PlatCon-18).

[3] Z. Shaik, and V. Asari. 2007. A robust method for multiple face tracking using Kalman filter. Applied Imagery Pattern Recognition Workshop. 36th IEEE.

[4] P. Viola, and M. J. Jones. 2004. Robust Real-Time Face Detection. International Journal of Computer Vision, Vol. 57, no. 2, 137-54

[5] R. Belaroussi, L. Prevost, and M. Milgram. 2005. Combining model-based classifiers for face localization. Ninth IAPR Conference on Machine Vision Applications. 290-293

[6] K.-S. Kim, D.-S. Jang, and H.-I. Choi. 2007. Real time face tracking with pyramidal Lucas-Kanade feature tracker. International Conference on Computational Science and Its Applications.

[7] Y. C. Wettum. 2017. Facial landmark tracking on a mobile device. BS thesis. University of Twente.

[8] B. D. Lucas, and T. Kanade. 1981. An Iterative Image Registration Technique with an Application to Stereo Vision. Proceedings of the 7th International Joint Conference on Artificial Intelligence. Vol. 2. 674-679

[9] S. Hare, S. Golodetz, A. Saffari, V. Vineet, M. M. Cheng, S. L. Hicks, and P. H. Torr. 2016. Struck: Structured Output Tracking with Kernels. IEEE Transactions on Pattern Analysis and Machine Intelligence. 38.10. 2096-2109

[10] Danelljan, Martin, et al. "Discriminative scale space tracking." IEEE transactions on pattern analysis and machine intelligence 39.8 (2017): 1561-1575.

[11] J. F. Henriques, R. Caseiro, P. Martins, and J. Batista. 2015. High-Speed Tracking with Kernelized Correlation Filters. IEEE Transactions on Pattern Analysis and Machine Intelligence. 37.3. 583-596

[12] D. E. King. 2009. Dlib-ml: A Machine Learning Toolkit. Journal of Machine Learning Research. 10. 1755-1758.

[13] DEST v.08 software library. 2016. http://github.com/cheind/dest/releases (visited on 15-12-16)

[14] K. Zhang, Z. Zhang, Z. Li, and Y. Qiao. 2016. Joint face detection and alignment using multitask cascaded convolutional networks. IEEE Signal Processing Letters. 23.10. 1499-1503

[15] R. Girshick, J. Donahue, T. Darrell, and J. Malik. 2014. Rich feature hierarchies for accurate object detection and semantic segmentation. Proceedings of the IEEE conference on computer vision and pattern recognition.

[16] R. Girshick. 2015. Fast r-cnn. ArXiv preprint arXiv:1504.08083

[17] S. Ren, K. He, R. Girshick, and J. Sun, 2015. Faster r-cnn: Towards real-time object detection with region proposal networks. Advances in neural information processing system.

[18] J. Redmon, S. Divvala, R. Girshick, and A. Farhadi. 2016. You only look once: Unified, real-time object detection. The IEEE conference on computer vision and pattern recognition.

[19] J. Long, E. Shelhamer, and T. Darrell. 2015. Fully convolutional networks for semantic segmentation. CVPR. pp. 3431-3440.

[20] H. Noh, S. Hong, and B. Han. 2015. Learning deconvolution network for semantic segmentation. ICCV, pp. 1520-1528.

[21] O. Ronneberger, P. Fischer, and T. Brox. 2015. U-net: Convolutional networks for biomedical image segmentation. MICCAI. pp. 234-241.

[22] V. Badrinarayanan, A. Kendall, and R. Cipolla. 2017. Segnet: A deep convolutional encoder-decoder architecture for image segmentation. IEEE transactions on pattern analysis and machine intelligence. 39.12. pp. 2481-2495.

[23] W. Y. Lo, and S. M. Puchalski. 2008. Digital image processing. Veterinary Radiology & Ultrasound 49 s1.

[24] J. Canny. 1986. A Computational Approach to Edge Detection. IEEE Transactions on Pattern Analysis and Machine Intelligence. Vol. 8. Pp. 679-698.

5
[25] C. Tomasi, and T. Kanade. 1991. Detection and Tracking of Point Features. Carnegie Mellon University Technical Report CMU-CS-91-132.